\documentclass[format=sigconf]{acmart}
\usepackage{amsmath}
\usepackage{courier}
\usepackage{dsfont}
\usepackage{booktabs}
\usepackage{multirow}
\usepackage{graphicx}
\usepackage{makecell}
\usepackage{subcaption}
\usepackage{enumitem}
\usepackage{hyperref}

\usepackage[algo2e,ruled,vlined,linesnumbered]{algorithm2e}
\usepackage{xspace}

\AtBeginDocument{%
  }

\copyrightyear{2026}
\acmYear{2026}
\setcopyright{cc}
\setcctype{by-nc-nd}
\acmConference[GECCO '26]{Genetic and Evolutionary Computation Conference}{July 13--17, 2026}{San Jose, Costa Rica}
\acmBooktitle{Genetic and Evolutionary Computation Conference (GECCO '26), July 13--17, 2026, San Jose, Costa Rica}
\acmDOI{10.1145/3795095.3805071}
\acmISBN{979-8-4007-2487-9/2026/07}

\begin{document}

\title{Block-Bench: A Framework for Controllable and Transparent Discrete Optimization Benchmarking}

\author{Furong Ye}
\email{f.ye@liacs.leidenuniv.nl}
\orcid{0000-0002-8707-4189}
\affiliation{%
  \institution{LIACS, Leiden University}
  \city{Leiden}
  \country{the Netherlands}
}

\author{Frank Neumann}
\email{frank.neumann@adelaide.edu.au}
\orcid{0000-0002-2721-3618}
\affiliation{%
  \institution{Optimisation and Logistics, 
  School of Computer Science and Information Technology, Adelaide University}
  \city{Adelaide}
  \country{Australia}
}

\author{Thomas B\"ack}
\email{t.h.w.baeck@liacs.leidenuniv.nl}
\orcid{0000-0001-6768-1478}
\author{Niki van Stein}
\email{n.van.stein@liacs.leidenuniv.nl}
\orcid{0000-0002-0013-7969}
\affiliation{%
  \institution{LIACS, Leiden University}
  \city{Leiden}
  \country{the Netherlands}
}

\newcommand{\OMM}{\textsc{OneMinMax}}
\newcommand{\LOTZ}{\textsc{Leading\-Ones-Trai\-ling\-Zeros}}
\newcommand{\OJZJ}{\textsc{OneJump-ZeroJump}}
\newcommand{\ORZR}{\textsc{One\-Royal\-Road-Zero\-Royal\-Road}}
\newcommand{\OM}{\textsc{OneMax}}
\newcommand{\LO}{\textsc{LeadingOnes}}
\newcommand{\Jump}{\textsc{Jump}}
\newcommand{\Eps}{\textsc{Epistasis}}
\newcommand{\NSGA}{{NSGA-II}}
\newcommand{\MOEAD}{{MOEA/D}}
\newcommand{\SMS}{{SMSEMOA}}
\newcommand{\tworate}{$(1+\lambda)$ \emph{two-rate} EA}
\newcommand{\var}{$(1+\lambda)$ \emph{var} EA}
\newcommand{\llGA}{$(1+(\lambda,\lambda))$ GA}
\newcommand{\rev}[1]{{#1}}

\begin{abstract}
We present a novel approach for constructing discrete optimization benchmarks that enables fine-grained control over problem properties, and such benchmarks can facilitate analyzing discrete algorithm behaviors.
We build benchmark problems based on a set of block functions, where each block function maps a subset of variables to a real value. Problems are instantiated through a set of block functions, weight factors, and an adjacency graph representing the dependency among the block functions. Through analyzing intermediate block values, our framework allows to analyze algorithm behavior not only in the objective space but also at the level of variable representations in the obtained solutions. This capacity is particularly useful for analyzing discrete heuristics in large-scale multi-modal problems, thereby enhancing the practical relevance of benchmark studies. 
We demonstrate how the proposed approach can inspire the related work in self-adaptation and diversity control in evolutionary algorithms.
Moreover, we explain that the proposed benchmark design enables explicit control over problem properties, supporting research in broader domains such as dynamic algorithm configuration and multi-objective optimization.
\end{abstract}

\maketitle
\section{Introduction}
Understanding the behavior of evolutionary algorithms (EAs) and iterative heuristics is important for enhancing the explainability, robustness, and novel design of algorithms~\cite{bartz2020benchmarking}. Recent advances in theory and benchmarking studies have enhanced fundamental analysis for iterative algorithms. Theoretical runtime analysis helps us to know the performance guarantee and the role of operators in searching for better solutions~\cite{doerr2019theory}. On the other hand, benchmark studies provide empirical evidence of algorithms' behavior across various problems~\cite{rapin2019exploring,doerr2020benchmarking,hansen2022anytime,marty2023benchmarking,van2025explainable}, and the numerous experimental datasets generated support advances in ML (machine learning)-driven automated algorithm selection or assembly methods for complex practical applications~\cite{vermetten2019online,biedenkapp2022theory,ye2022automated,adriaensen2022automated}.

However, in contrast to the rapid development of systematic analyses for continuous optimization, related work in discrete optimization is either limited to classic functions in theoretical analysis~\cite{doerr2019theory} or tailored to particular application domains such as satisfiability (SAT)~\cite{satcompetition}, routing~\cite{uchoa2017new}, scheduling~\cite{vallada2015new}, etc. These practical problems often come with high dimensionality, and the objective search space is too complex to support effective analysis of algorithm behavior. For example, escaping local optima is usually considered an important capacity of algorithms, but this is evident only from the final results, without illustrating the frequency and timing of encounters with these local optima~\cite{luo2014ccls}. 
Existing general discrete benchmarks~\cite{nevergrad,doerr2020benchmarking} evaluate algorithms primarily based on the fitness values following the concept of black-box optimization. However, this work largely overlooks the impact of variable representations, even though many operators have been specifically proposed to address specific representation patterns in practical domains~\cite{helsgaun2000effective}.
Some initial effort has been conducted to implement tunable features~\cite{weise2018difficult} for discrete benchmarks, but their usage remains rather limited.
The lack of granular and representation-aware discrete benchmarks hinders practical adoption and communication between theoretical and practical research, as well as prevents effective idea exchange across practical scenarios. Consequently, we lack a comprehensive foundation for examining and promoting advanced techniques, such as parameter control and automated algorithm design, in general discrete optimization studies. 

We propose block-structured optimization benchmarks to effectively address the issues faced by the discrete community. First, by introducing a set of \emph{block functions} to form objective functions, our benchmark problems enable analysis of the sensitivity of algorithms to different local optima patterns by observing corresponding block values. The block structure is helpful for analyzing algorithms' behavior at the level of variable representations with determined settings of block functions. It also benefits the study of large-scale discrete optimization, as we can analyze block values in a reduced-dimensional space. In addition, by controlling the contribution (i.e., weights) of each block function to the objective function and its dependencies, we can formulate various benchmark problems with deterministic landscapes and local optima, revealing the potential of block-structured benchmarks in building a foundation for examining broader algorithm design topics in discrete optimization.

Overall, the contributions of this paper are:

\begin{itemize}[topsep=0pt]
    \item Proposing a novel approach to construct discrete optimization benchmarks. This approach enables controlling problem properties, particularly the objective search space and local optima patterns. In addition, it supports analyzing large-scale discrete optimization over block values in a lower-dimensional space.
    \item Benefiting from the block-structures, we examine several questions that existing benchmarks cannot address.
    \begin{itemize}[leftmargin=*]
        \item We analyze several self-adaptive EAs on problems with diverse landscape properties, highlighting the challenges posed by dynamic landscapes.
        \item We revisited the idea that maintaining diversity can help EAs deal with multi-modality, showing that this advantage varies with modality and landscape complexity.
        \item With our bi-objective optimization benchmarks, our experimental results reveal that simple evolutionary multi-objective optimizers (SEMOs) and \MOEAD~encounter greater difficulty in handling condition based landscapes, compared to \NSGA~and \SMS.
    \end{itemize}
    \item We provide a list of discrete benchmarks and together with corresponding experimental results in \rev{our repository\footnote{\href{https://github.com/FurongYe/BlockBuildingBenchmark}{https://github.com/FurongYe/BlockBuildingBenchmark}}.}
\end{itemize}

\section{Related Work}

In this section, we provide a brief overview of the state of the art in benchmarking in Evolutionary Computation, and summarize the key discrete Evolutionary Algorithms used in our work.

\subsection{Evolutionary Computation Benchmarking}
To better understand the behavior and performance guarantees of evolutionary computation (EC), theoretical and empirical analyses~\cite{doerr2019theory} have been widely studied in recent years. Meanwhile, parameter tuning and control~\cite{eiben2002parameter}, and other learning-based techniques~\cite{liu2023learning,chen2023using} have been proposed to improve the efficiency of algorithms. To support these studies and ensure fairness and reproducibility of the results, several benchmarking platforms have been developed to generate standard experimental data~\cite{coco,nevergrad,ioh}.  
\subsubsection{Practice in the continuous and discrete optimization}
Continuous optimization benchmarks have been developed rapidly in recent years. The well-known bbob suite~\cite{bbob} has significantly enhanced related studies in the continuous optimization domain. Its problems are systematically classified based on well-defined properties such as multi-modality, global structure, separability, etc~\cite{bbob,mersmann2011exploratory}. Apart from the original COCO platform~\cite{coco}, the bbob suite has also been integrated into other platforms such as IOHprofiler~\cite{ioh} and Nevergrad~\cite {nevergrad}. An extended many-affine combinations of existing problems have been proposed to generate more diverse problems~\cite{vermetten2025ma}, and Exploratory Landscape Analysis has also been proposed to investigate the properties of unknown problems~\cite{mersmann2011exploratory}. Moreover, ML-related learning techniques have been studied, addressing topics such as algorithm selection for continuous optimization~\cite{vermetten2019online,kostovska2022per}. 

While the design of continuous benchmarks is driven primarily by capturing different problem characteristics to support benchmarking and a broad range of applications, discrete benchmarks, which are usually \rev{constructed with simple but explainable landscapes}, come from runtime analysis. The pseudo-Boolean optimization (PBO) benchmark set proposed in IOHprofiler~\cite{doerr2020benchmarking} consists of classic theory-oriented problems and practical constrained problems with pre-defined penalty factors.
Another common approach is to discretize continuous benchmark sets, such as those from bbob and the CEC competitions~\cite{bbob}, to evaluate discrete EC methods. Nevertheless, these discretized problems can not exactly capture the intrinsic difficulties of discrete landscapes. 

Beyond these generic benchmarks, many discrete benchmark suites are tailored to specific research domains. For example, theoretical studies commonly focus on the classic \OM, \LO, \Jump, etc., which are characterized by particular and understood landscape properties. In contrast, practical research directly works on domain-specific test suites such as those for the SAT~\cite{satcompetition} problem, the vehicle routing problem~\cite{uchoa2017new}, and flowshop scheduling~\cite{vallada2015new}. While these benchmarks are highly relevant to practical situations, the corresponding algorithm performance is usually evaluated solely on the obtained solution quality, and the underlying search spaces are often too complex to provide interpretable insights into algorithm behavior. Moreover, benchmarking and theoretical studies have been developed to address specific problem classes, such as submodular problems~\cite{neumann2023benchmarking,friedrich2015maximizing}. 

\subsection{Discrete Evolutionary Algorithms}
In this section, we provide an overview of several algorithms considered in this work, and more details are described in the Appendix.
Note that we particularly work on the 
pseudo-Boolean optimization (PBO) problems $f: \{1,0\}^n \rightarrow \mathds{R}^d$ in this paper and show examples illustrating both the single- and multi-objective case.

\subsubsection{Self-adaptation in Single-objective Optimization}
\label{sec:self}
The $(1+\lambda)$ EAs apply a mutation-only scheme that applies the current best solution as the parent and creates $\lambda$ offspring solutions at each iteration. The offspring is generated by flipping $\ell$ distinct bits of the parent, and the value $\ell$ is sampled by a distribution correlated to the mutation rate $p$. We consider the following $(1+\lambda)~$EAs that follow different methods of sampling $\ell$.
\begin{itemize}[leftmargin=*]
    \item $(1+1)~$EA with a fixed mutation rate $p = 1/n$.
    \item Fast Genetic Algorithm (fGA)~\cite{fga}, which follows the $(1+1)$ scheme and samples $\ell$ from a fixed long-tail distribution.
    \item \tworate~\cite{tworate}, of which $p$ is adapted by $2p$ or $p/2$ based on a success rule at each iteration.
    \item \var~\cite{varEA}, which sampling $\ell$ from a normal distribution with adaptive mean and variance.
\end{itemize}
Additionally, we consider another theory-inspired self-adaptive algorithm applying both crossover and mutation:
\begin{itemize}[leftmargin=*]
    \item \llGA~\cite{doerr2018optimal}, which adapts the values of $\lambda$, and this $\lambda$ determines its crossover and mutation rate dynamically.
\end{itemize}
We set $\lambda = 10$ throughout this paper, and refer readers to the recent benchmark study in~\cite{doerr2020benchmarking} for their performance discussions.

\rev{We note that the algorithms listed above are primarily theory-driven. While such algorithms have been widely studied and evaluated on established benchmarks~\cite{doerr2020benchmarking,nevergrad}, other advanced techniques, such as partition crossover and linkage learning, have been proposed to address the complexity of practical problems~\cite{thierens2010linkage,tinos2015partition,bosman2016expanding}. Furthermore, tailored local search methods have been developed for specific applications, including the maximal satisfiability problem and constrained pseudo-boolean optimization~\cite{lei2021efficient,chu2024enhancing}.}

\subsubsection{Bi-objective Optimization} 
\label{sec:BIO}
We refer readers to~\cite{deb2016multi} for the essential concepts of multi-objective optimization. In this work, we particularly consider five algorithms, SEMO, Global SEMO (GSEMO)~\cite{antipov2023rigorous}, \NSGA~\cite{nsga2}, \MOEAD~\cite{moead}, and \SMS~\cite{sms}. In our experiments, the pymoo~\cite{pymoo} package is used for the latter three algorithms, with population size $n$, and we set the mutation rate of GSEMO to $p=1/n$. The computational budget is set to $n^3$.

Theoretical runtime analyses~\cite{jansen2013analyzing,neumann2010bioinspired,doerr2019theory}
have deepened our understanding of the performance of evolutionary algorithms and provided valuable insights into algorithm design for multi-objective problems~\cite{HorobaN08,HorNeuFOGA09,DBLP:series/sci/HorobaN10}. 
Extensive runtime analyses have been conducted on functions such as \OMM, \LOTZ, \OJZJ, and \ORZR~\cite{zheng2022first,giel2006effect,antipov2023rigorous,laumanns2004running,doerr2016runtime,doerr2024proven,nguyen2015population}.
Theoretical studies have also addressed combinatorial problems such as multi-objective minimum spanning tree ~\cite{neumann2007expected,cerf2023first,do2023rigorous}, shortest paths~\cite{DBLP:journals/ec/Horoba10}, and submodular problems~\cite{qian2019maximizing,badanidiyuru2014fast,neumann2020optimising,friedrich2015maximizing}. A recent study~\cite{liang2025problem} introduced an extended benchmark set by combining different objective formulations of these problems and illustrated several of their properties, but the benchmarks remain restricted to specific landscape properties.

\section{Block-Structured Benchmarks}
We introduce a novel approach to construct discrete optimization benchmarks that allows explicit control over problem properties and provides additional insights into algorithmic behavior. \rev{\emph{Block} is not a novel concept in the EC community. It has appeared in classic building block hypothesis~\cite{holland1992adaptation}, as well as in recent theoretical analyses~\cite{zheng2024runtime}. Moreover, some advanced learning techniques aim to identify representative blocks that capture specific variable interactions~\cite {thierens2010linkage,tinos2015partition}. The well-known NK-landscape problem can also be viewed as being constructed from a set of interacting blocks~\cite{altenberg1996b2}. In contrast, this paper focuses on benchmark construction, aiming to address the challenges of large-dimensional discrete search spaces by introducing explainable units (i.e., blocks) and mitigating the bottlenecks introduced by randomness.}

In this paper, we focus on PBO, but the proposed approach can naturally extend to general discrete search spaces.
Formally, we define a pseudo-Boolean problem as a function $f: \{0,1\}^n \rightarrow \mathds{R}, x \mapsto f(x)$,
which is composed of a set of block functions $\{v_1,\ldots, v_m\}$, and we work on \emph{maximization}. Each block function $v_i: \{0,1\}^{n_i} \rightarrow \mathds{R}_{\ge 0}, x^i \mapsto v_i(x^i)$ operates on a substring $x^i$ of $x$ with length $n_i = |x^i| $, where $i\in [m]$.  We denote $\{1,\ldots,m\}$ by $[m]$  throughout this paper.
An instance of problem $f$ is specified by the set of block values $V = \{v_1(x^1), \ldots, v_m(x^m) \}$\footnote{We denote $x$ is partitioned into $m$ equal-length substrings $x^1, \ldots,x^m$ in this paper.}, a corresponding set of weights for the blocks $W =\{w_1,\ldots,w_m\}$, a dependency graph $G = (V,E)$ defining the dependencies among block functions by an adjacency matrix $E \in \mathds{R}^{m\times m}$, and a constant vector $A = \{a_1,\ldots,a_m\}$. According to the structure of the dependency graph $G$, we define two categories of benchmarks. Detailed descriptions of a selected benchmark set, following the definition below, are provided in Table~\ref{tab:problems}.

\begin{table*}[htb]
    \centering
    \small
    \begin{tabular}{c|c|c|c|c|c|c}
        \toprule
        & \textbf{Type} & \textbf{Block functions} & $A$ & \textbf{$W$} & \textbf{ $E$} & \textbf{$B$}\\
        \midrule

          F1  &  DBP  &\OM& $\{0\}^m$ & $\{1\}^m$ & $\{0\}^{m \times m}$ & - \\
         \hline
         F2  &  DBP  &\LO& $\{0\}^m$ & $\{1\}^m$ & $\{0\}^{m \times m}$ & - \\
         \hline
         F3  &  DBP  & \textsc{Jump}$_k$& $\{0\}^m$ & $\{1\}^m$ & $\{0\}^{m \times m}$ & - \\
             \hline

         F4  &  DBP  &\Eps& $\{0\}^m$ & $\{1\}^m$ & $\{0\}^{m \times m}$ & - \\
         \hline
        
         F5  &  DBP  & \makecell{\textsc{OneMax},\textsc{LeadingOnes}, \\ \textsc{Jump}$_3$,\textsc{Epistasis}}& $\{0\}^m$ & $\{1\}^m$ & $\{0\}^{m \times m}$ & - \\
         \hline

         F6  &  DBP  & \makecell{\textsc{OneMax},\textsc{Jump}$_2$, \\ \textsc{Jump}$_3$,\textsc{Epistasis}}& $\{0\}^m$ & $\{1\}^m$ & 
         $\{0\}^{m \times m}$  & - \\
         \hline

         F7  &   GCP  & \textsc{Jump}$_3$& $\{0\}^m$ & $\{1\}^m$ & \makecell{$e_{ij} = 1$, if $j = i+1$; \\  $e_{ij} = 0$, otherwise} & $\{\frac{n}{m} +3\}^m$\\ 
         \hline
        F8  &   GCP  &\Eps& $\{0\}^m$ & $\{1\}^m$ & \makecell{$e_{ij} = 1$, if $j = i+1$; \\  $e_{ij} = 0$, otherwise} & $\{\frac{n}{m}\}^m$\\ 
         \hline

        F9 &   GCP  & \makecell{\textsc{OneMax},\textsc{Jump}$_2$, \\ \textsc{Jump}$_3$,\textsc{Epistasis}}& $\{0\}^m$ & $\{1\}^m$ & \makecell{$e_{ij} = 1$, if $j = i+1$; \\  $e_{ij} = 0$, otherwise}  & $\{\frac{n}{m}+5\}^m$\\
         \hline

         F10 &   GCP  & \makecell{\textsc{OneMax},\textsc{LeadingOnes}, \\ \textsc{Jump},\textsc{Epistasis}}& $\{0\}^m$ & $\{1\}^m$ & \makecell{$e_{ij} = 1$, if $j = i+1$; \\  $e_{ij} = 0$, otherwise}  & $\{\frac{n}{m} + 3\}^m$\\
         \midrule

         \multirow{2}{*}{BF1}  &  DBP  & \textsc{OneMax}& $\{0\}^m$ & $\{1\}^m$ & $\{0\}^{m \times m}$ & - \\ 
         
           &  DBP  & \textsc{OneMax}& $\{a_i = n_i\}$ & $\{-1\}^m$ & $\{0\}^{m \times m}$ & -  \\ 
         \hline 
           \multirow{2}{*}{BF2}  & GCP  & \textsc{OneMax} & $\{0\}^m$ & $\{1\}^m$ & \makecell{$e_{ij} = 1$, if $j = i+1$; \\  $e_{ij} = 0$, otherwise}  & $\{b_i = n_i\}$ \\
          & GCP  & \textsc{OneMax}& $\{a_i = n_i\}$ & $\{-1\}^m$ & \makecell{$e_{ij} = 1$, if $j = i-1$; \\  $e_{ij} = 0$, otherwise} & $\{0\}^m$\\ 
            \hline
             \multirow{2}{*}{BF3}  & GCP  &\LO & $\{0\}^m$ & $\{1\}^m$ & \makecell{$e_{ij} = 1$, if $j = i+1$; \\  $e_{ij} = 0$, otherwise}  & $\{b_i = n_i\}$ \\
          & GCP  & \LO& $\{a_i = n_i\}$ & $\{-1\}^m$ & \makecell{$e_{ij} = 1$, if $j = i-1$; \\  $e_{ij} = 0$, otherwise} & $\{0\}^m$\\ 
            \hline
           \multirow{2}{*}{BF4}  & GCP  & \OM & $\{a_i | a_i =\frac{n_i(1-w_i)}{2}\}$ & $\{w_i|w_i = \delta_{i\%5}, \delta =(1,-1,1,1)\}$ & \makecell{$e_{ij} = 1$, if $j = i+1$; \\  $e_{ij} = 0$, otherwise}  &  $\{b_i | b_i =\frac{n_i(1+w_i)}{2}\}$  \\
          & GCP  & \OM & $\{a_i | a_i =\frac{n_i(1-w_i)}{2}\}$ & $\{w_i|w_i = \delta_{i\%5}, p =(-1,1,1,-1)\}$ & \makecell{$e_{ij} = 1$, if $j = i-1$; \\  $e_{ij} = 0$, otherwise}  &  $\{b_i | b_i =\frac{n_i(1+w_i)}{2}\}$ \\
         \hline
        \multirow{2}{*}{BF5}  & GCP  & \LO & $\{a_i | a_i =\frac{n_i(1-w_i)}{2}\}$ & $\{w_i|w_i = \delta_{i\%5}, \delta =(1,-1,1,1)\}$ & \makecell{$e_{ij} = 1$, if $j = i+1$; \\  $e_{ij} = 0$, otherwise}  &  $\{b_i | b_i =\frac{n_i(1+w_i)}{2}\}$  \\
          & GCP  & \LO & $\{a_i | a_i =\frac{n_i(1-w_i)}{2}\}$ & $\{w_i|w_i = \delta_{i\%5}, \delta =(-1,1,1,-1)\}$ & \makecell{$e_{ij} = 1$, if $j = i-1$; \\  $e_{ij} = 0$, otherwise}  &  $\{b_i | b_i =\frac{n_i(1+w_i)}{2}\}$ \\
         \bottomrule
    \end{tabular}
    \caption{The instances of block-structure problems mentioned in this paper. $n$ is the problem dimensionality, $m$ is the number of blocks, $i,j \in [m]$. The functions listed in the ``Blocks functions'' column are sequentially added as $m$ increases. ``F*'' are single-objective problems, and ``BF*'' are bi-objective problems formed by the corresponding two problem instances.}
    \label{tab:problems}
\end{table*}

\subsection{Undirected Graph Based Problems}

\begin{figure}[tb!]
    \includegraphics[trim=8 8 5 5, clip,width=0.3\linewidth]{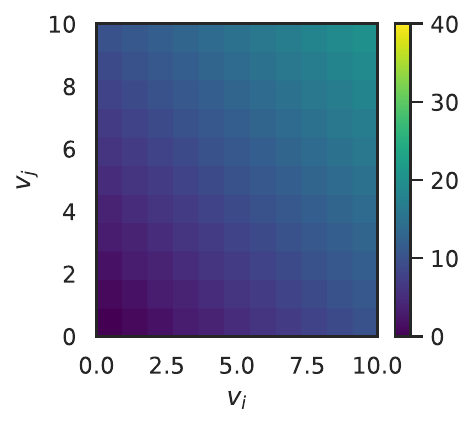}
     \includegraphics[trim=8 8 5 5, clip,width=0.3\linewidth]{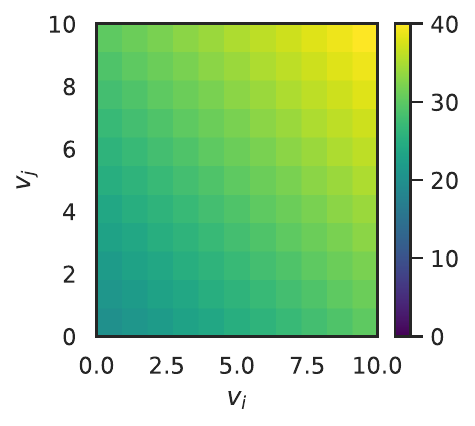}
    \caption{Correlations between block function values (in axes) to the minimal (\emph{Left}) and maximal (\emph{Right}) attainable objective values (indicated by colors) for a DBP.}
    \label{fig:DBP}
\end{figure}

\begin{figure}[tb!]
    \includegraphics[trim=8 5 5 0, clip,width=0.3\linewidth]{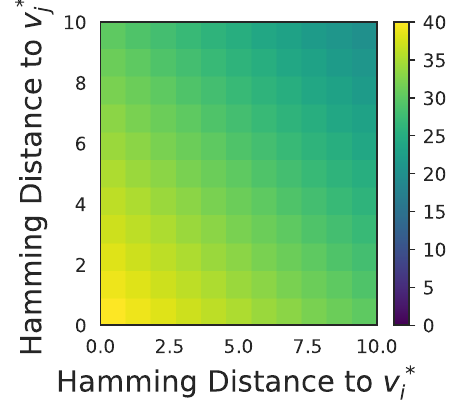}
    \includegraphics[trim=8 5 5 0, clip,width=0.3\linewidth]{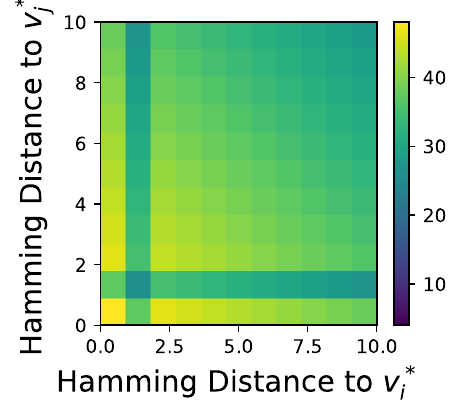}
    \includegraphics[trim=8 5 5 0, clip,width=0.3\linewidth]{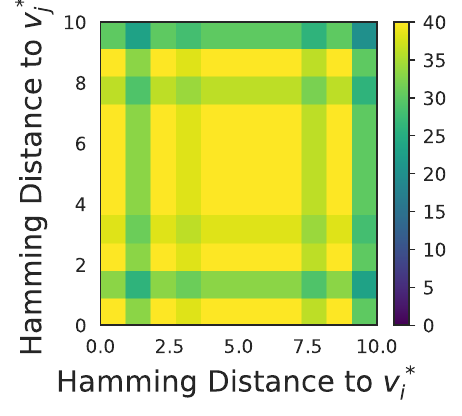}
    \caption{Diverse landscapes constructed by block functions \OM (Left, F1), \textsc{Jump}$_2$ (Mid, F3), or \Eps~(Right, F4).}
    \label{fig:DBP-d}
\end{figure}
A general representation of our dependency-based problems (DBPs) is defined by 
\begin{equation}
    f(x) =  \underset{i\in [m]}{\sum} (a_i + w_i v_i(x^i)) + \underset{{i,j \in [m]}}{\sum} e_{ij} v_i(x^i) v_j(x^j) 
\label{eq:ug}
\end{equation}
where $v_i(x^i)$ denotes the value of the $i$-th block function.
We use an \emph{undirected graph} $G = (V,E)$ to represent dependencies among block functions, in which the node set is denoted by the set of block functions $V$. To avoid redundant calculations, the edge set is represented by a \emph{strict lower triangular matrix} $E$. $e_{ij} = 1$ indicates dependency exists between $v_i$ and $v_j$,  $e_{ij} = 0$ otherwise. As shown in Eq.~\ref{eq:ug}, given the weight vector $W$, the dependency matrix $E$, and a constant vector $A$, the mapping from block values $V$ to the objective $f$ is fully determined. Furthermore, we construct specific landscape structures by controlling the definitions of block functions. 

As an example, Figure~\ref{fig:DBP} illustrates the correlation between block values for $40$-dimensional (40D) DBPs 
with $m=4$ block functions $v(\cdot) \in \{0,\ldots,10\}$. In the plotted setting, all blocks contribute equally to the objective, i,e., $w_i = 1, \forall i \in [m]$, all block functions are independent from each other, i.e., $e_{ij} = 0$ and $a_i = 0, \forall i,j\in [m]$.

Building upon this setting, Figure~\ref{fig:DBP-d} demonstrates how different block functions determine the structure of the search landscape. With the above setting plotted in Figure~\ref{fig:DBP}, we consider three types of block functions: \OM, \Jump, and a modified \OM~with \Eps, defined as below:
\begin{equation}
    \OM : v(x) = \sum_{i \in [n]} x_i,
\end{equation}
\begin{equation}
    \textsc{Jump}_k : v(x) =
\left\{
\begin{array}{ll}
|x|_1 + k, & \text{if } |x|_1 \le n- k \text{ or } |x|_1 = n  \\
n - |x|_1, & \text{otherwise}
\end{array}
\right. ,
\end{equation}
where $|x|_1 = \sum_{i \in [n]} x_i$.
\begin{equation}
    \Eps : v(x) = \OM(x'),
\end{equation}
where $x'$ is transformed from $x$ by mapping its Hamming-1 neighbors $s_1,s_2,\in \{0,1\}^\nu$ to strings 
with distance at least $\nu-1$. We set $\nu = 3$ in this paper.
Details of the transformation refer to~\cite{weise2018difficult}.

\begin{figure}[tb!]
    \includegraphics[trim=8 5 5 5, clip,width=0.70\linewidth]{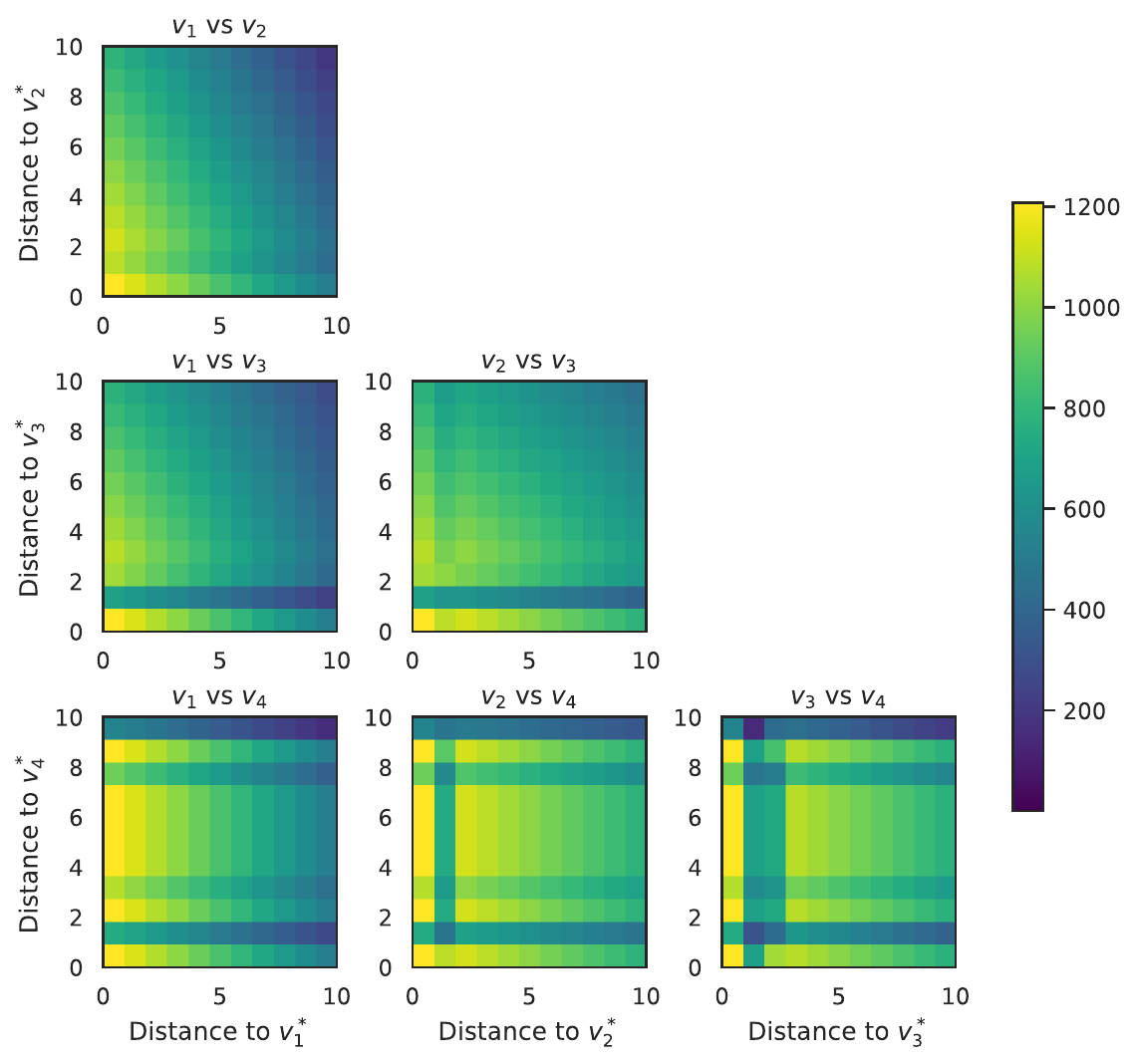}
\caption{A complex but designed landscape constructed by several distinct block functions.}
    \label{fig:DBP-mixed}
\end{figure}

Figure~\ref{fig:DBP-d} shows the best attainable objective value vs. the Hamming distance between each block’s substring $x^i$ and its corresponding optimum. Since all block functions are identical for the problems plotted in Figure~\ref{fig:DBP-d}, the correlation of each pair of block functions is the same. The \emph{Left} subfigure illustrates the identical search space of \OM, which exhibits a smooth and monotonic gradient toward the optimum.  The \emph{Mid} subfigure, using \Jump$_2$ block function, demonstrates a pronounced discontinuity near the optimum, while the majority of its search space remains smooth and monotonic as \OM. The \emph{Right} subfigure shows a more irregular and rugged landscape caused by the epistasis transformation. 
While these landscape properties have been well studied individually in existing benchmarks, our block-structured benchmark can facilitate the coexistence of heterogeneous landscape characteristics within a single objective function. Figure~\ref{fig:DBP-mixed} illustrates this capacity by presenting the best objective value attainable for a $40$D problem composed of \OM, \textsc{Jump}$_2$, \textsc{Jump}$_3$, and \textsc{Epistasis} (F6), regarding the Hamming distance between blocks' substrings and their corresponding optimum.
This example demonstrates our approach can construct complex landscapes with diverse but controllable properties.

\subsection{Directed Acyclic Graph Based Problems}
We define gate-constrained problems (GCP) by representing correlation among block functions using a \emph{directed acyclic graph} $G(V,E)$, in which $e_{ij} \neq 0, e_{ij} \in \{0,1\}$ indicates there exists a directed edge (i.e., gate constraint) from block $v_i$ to $v_j$. Let $\text{Anc}(i)$ denote the set of predecessors of block $v_i$, defined as all nodes $v_j$ for which there exists a path from $v_j$ to $v_i$. Formally, $\text{Anc}(i) = \underset{j \in \text{P}(i)}{\cup} (\text{Anc}(j) \cup j), \quad P(i) = \{j | e_{ji} \neq 0 , i,j \in [m]\}$. Then, a $n$-dimensional GCP with $m$ block functions is defined by
\begin{equation}
f(x) = \underset{{i\in [m]}}{\sum} (a_i + w_i v_i(x^i)) \cdot c_i,
\end{equation}
where the gate constraint function $c_i$ is given by
\begin{equation}
     c_i(x) = \underset{j \in \text{Anc}(i)}{\Pi} g(v_j(x^j),b_j), \quad g(v_j(x^j),b_j) = \mathds{1}_{v_j(x^j) \ge b_j}
\label{eq:gate}
\end{equation}

The contribution of each block function $v_i$ to the objective value $f(x)$ is thus controlled by a gate function $c_i$, which depends on the values of all blocks that have a directed path to $v_i$. According to Eq.~\ref{eq:gate}, block $v_i$ contributes to the objective $f$ only if all block values in $\text{Anc}(i)$ satisfy the specific constraints, e.g., reaching the optimum of their respective block function.
Overall, the weight vector $W$, the adjacency matrix $E$ of a directed acyclic graph, the constant vector $A$, the constraint factors $B = \{b_1,\ldots, b_m\}$, and the definition of block functions $V$ determine a GCP instance. For example, we can formulate a $n$-dimensional \LO~problem:
\begin{equation}
    \LO: v(x) = \sum_{i\in[n]}{\prod_{j\in[i]}{x_j}},
\end{equation}
using a GCP with a setting of $m = n$ \OM~block functions,
$E_1 = \left\{
\begin{array}{ll}
e_{ij} = 1, &\text{if } j = i+1 \\
0, & \text{otherwise}
\end{array}
\right.$, $a_i = 0$, and $b_i = 1, \forall i,j \in [m]$.

\begin{figure}[tb!]
    {\includegraphics[trim=8 8 5 5, clip,width=0.8\linewidth]{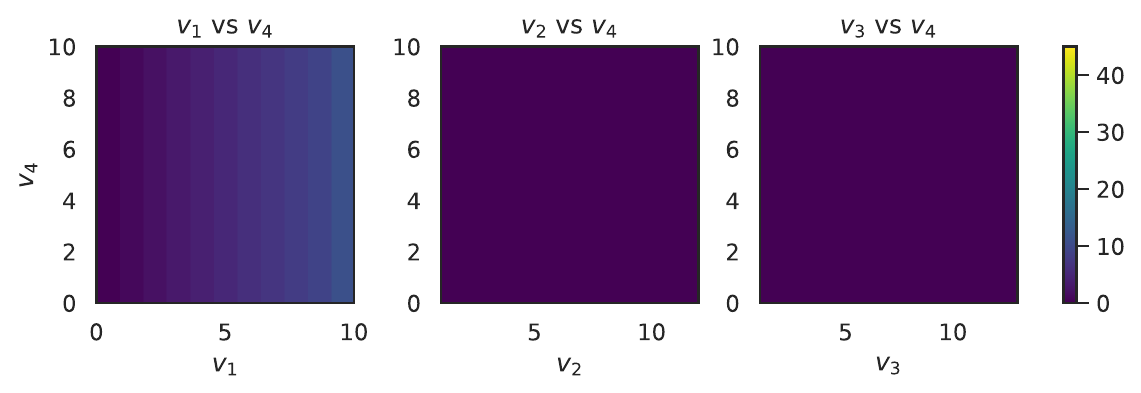}
    }
    {\includegraphics[trim=8 8 5 5, clip,width=0.8\linewidth]{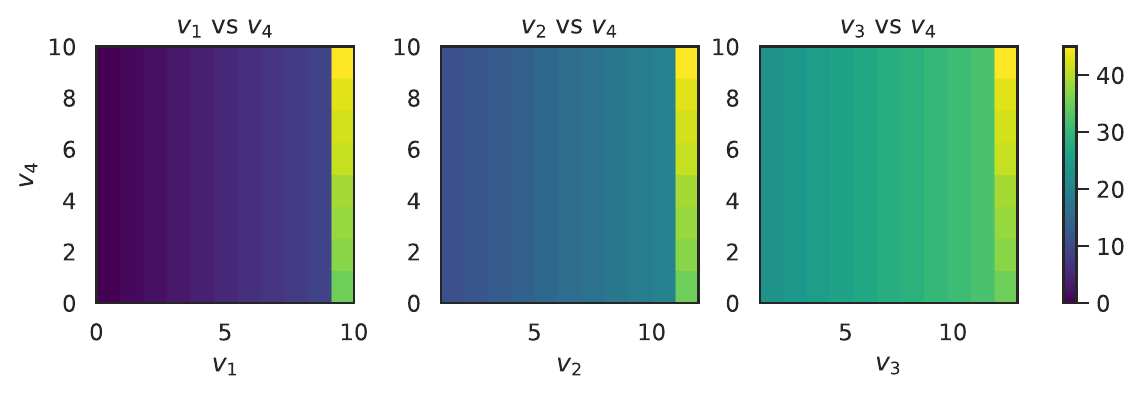}}
    \caption{Correlations between block function values (in axes) to the minimal (Top) and maximal (Bottom) attainable objective values (indicated by colors) for a GCP. We present the pairs between $v4$ and the others.
    }
    \label{fig:GCP}
\end{figure}

We plot in Figure~\ref{fig:GCP} the minimal and maximal attainable objective value of $40$D GCPs regarding the block values, with the setting of $m=4$ block functions $v(\cdot) \in \{0,\ldots,10\}$, $W = \{1\}^m$, $E = E_1$, $a_i = 0$ and $b_i = n_i, \forall i,j\in[m]$

\begin{figure}[tb!]
    \includegraphics[trim=8 5 5 5, clip,width=0.8\linewidth]{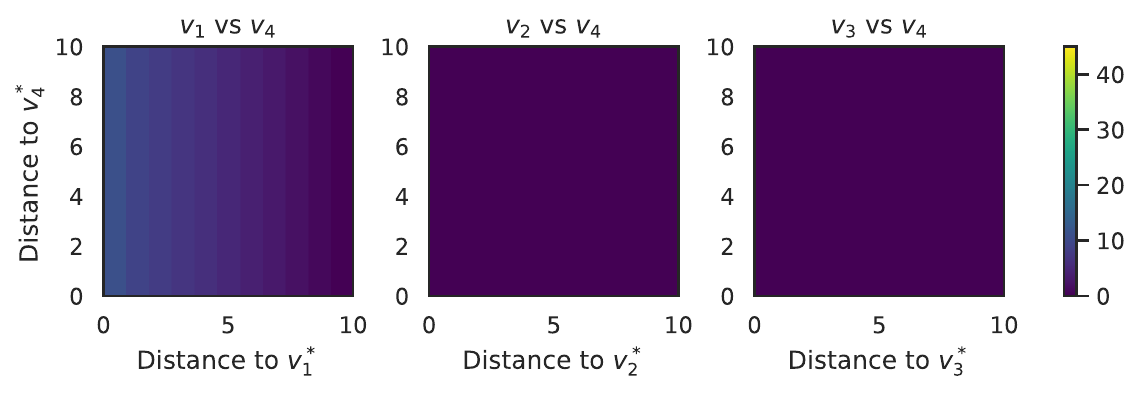}
    \includegraphics[trim=8 5 5 5, clip,width=0.8\linewidth]{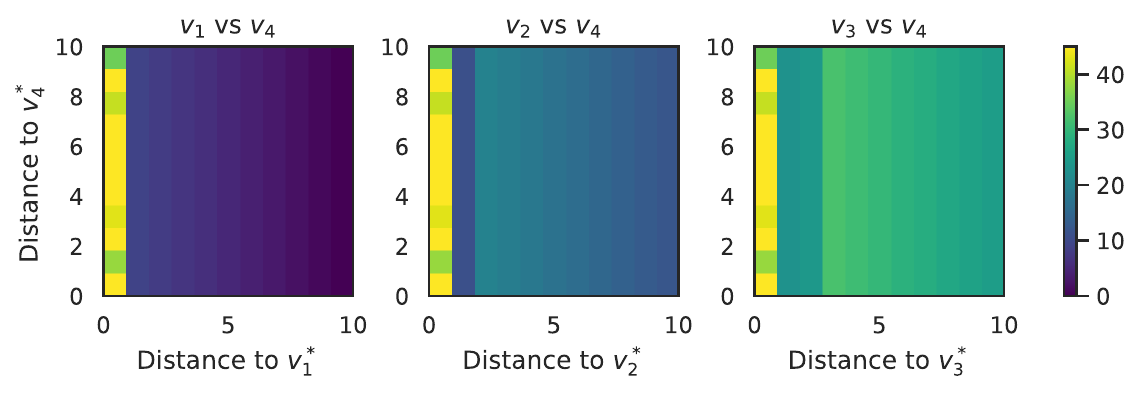}
    \caption{The minimal (Top) and maximal (Bottom) attainable objective values with respect to the corresponding Hamming distance of each block to its optimal solution, for the GCP F10 with the same objective space in Figure~\ref{fig:GCP}.}
    \label{fig:GCP-mixed}
\end{figure}

Although all block functions obtain identical weights (all 1s), the gate-constrained design introduces a strong asymmetry among different blocks. Figure~\ref{fig:GCP} shows large regions of neutrality in the lower bounds of attainable objective values, indicating that no meaningful objective gradient can be achieved when constraints are not satisfied. In contrast, the upper bounds of attainable objective values exhibit sharp fitness cliffs rather than gradual slopes in Figure~\ref{fig:DBP}. These neutrality landscape patterns and sharp fitness cliffs with respect to the Hamming distance to optimal blocks can also be observed in Figure~\ref{fig:GCP-mixed} (F10), in which a GCP is embedded with heterogeneous block functions. 
Recall that the definition of block functions introduces structured modality and local optima patterns into our benchmarks. For DBPs, the objective values typically change gradually as the block values are updated, since all blocks are simultaneously active and jointly contribute to the objective function. However, the contribution of block functions is hierarchically ordered for GCPs due to the gate constraints, resulting in ordered local optima patterns.

\section{Novel Insights from Benchmarking}
While our block-structured approach supports generating various benchmark problems, we illustrate how it can yield novel insights by using specific problem instances. Particularly, we address the topics of \emph{self-adaptation}, \emph{evolutionary multimodal optimization}, and \emph{bi-objective optimization} by revisiting several existing works.
\subsection{Self-Adaptation with Coexisting Landscape Properties}
\begin{figure}[tb!]
    \subcaptionbox{Convergence in terms of the best-found fitness.}{\includegraphics[trim=8 5 5 5, clip,width=0.9\linewidth]{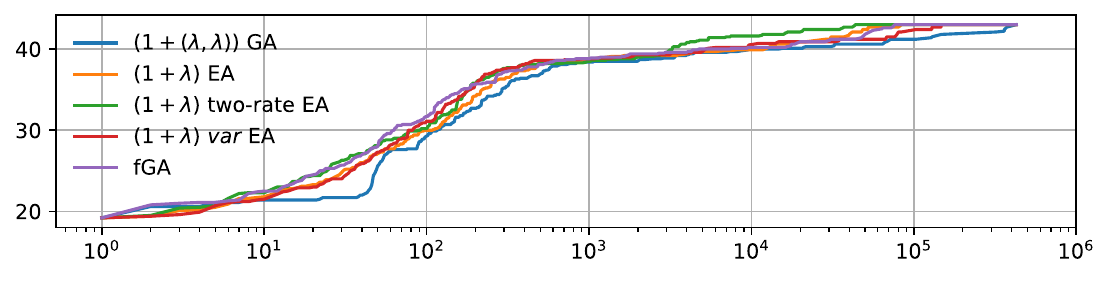}}
    \subcaptionbox{Block values of \OM, \LO, \textsc{Jump}$_3$, and \Eps.}{\includegraphics[trim=8 5 5 5, clip,width=0.22\linewidth]{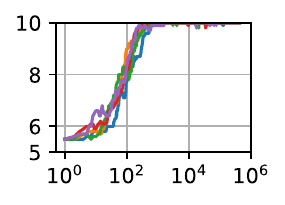}
    \includegraphics[trim=8 5 5 5, clip,width=0.22\linewidth]{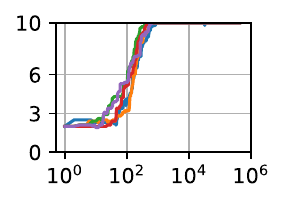}
    \includegraphics[trim=8 5 5 5, clip,width=0.22\linewidth]{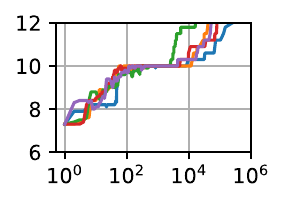}
    \includegraphics[trim=8 5 5 5, clip,width=0.22\linewidth]{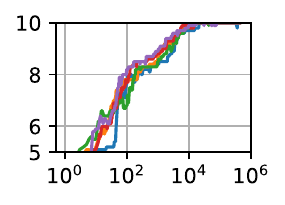}}
    \caption{Convergence process of EAs for F5 over the FEs. Results are the average of 50 runs.}
    \label{fig:SELF-UG}
\end{figure}

\begin{figure}[tb!]
    \subcaptionbox{Convergence in terms of the best-found fitness.}{\includegraphics[trim=8 5 5 5, clip,width=0.9\linewidth]{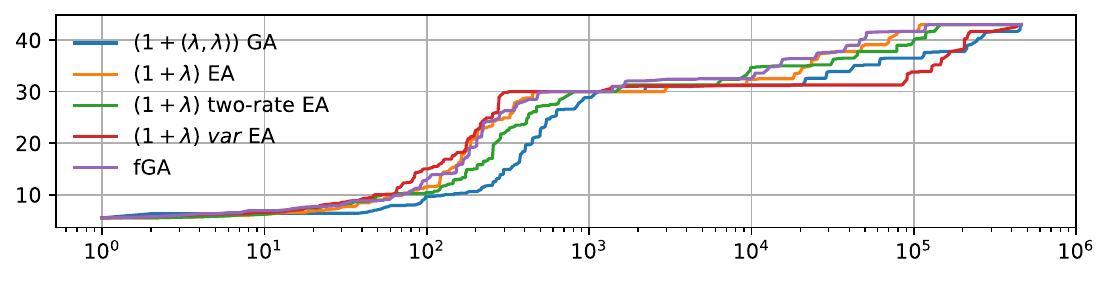}}
    \subcaptionbox{Block values of \OM, \LO, \textsc{Jump}$_3$, and \Eps.}{\includegraphics[trim=8 5 5 5, clip,width=0.22\linewidth]{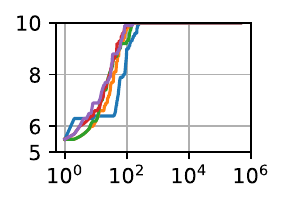}
    \includegraphics[trim=8 5 5 5, clip,width=0.22\linewidth]{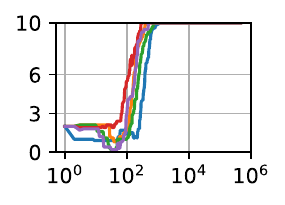}
    \includegraphics[trim=8 5 5 5, clip,width=0.22\linewidth]{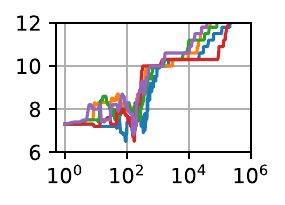}
    \includegraphics[trim=8 5 5 5, clip,width=0.22\linewidth]{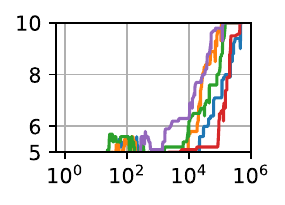}}
    \caption{Convergence process of EAs for F10 over the FEs. Results are the average of 50 runs.}
    \label{fig:SELF-DAG}
\end{figure}

Recall that the $(1+\lambda)~$EAs generate offspring by flipping $\ell$ bits of the parent solutions, and the mutation rate, which determines the sampling of $\ell$, has a vital impact on the performance of algorithms. Various self-adaptive mutation rate methods have been proposed (see Section~\ref{sec:self}). However, these methods have been commonly evaluated on particular problems such as \OM, \LO, and \Jump~\cite{doerr2018optimal,tworate,varEA,doerr2020benchmarking,fga}, without considering more complex landscapes. By leveraging the proposed benchmark, we are able to investigate \emph{how these self-adaptive methods perform for the coexistence of multiple landscape properties within a single search space}.

Precisely, we investigate the performance of the five algorithms introduced in Section~\ref{sec:self} on $40$D DBP and GCP instances with $m=4$ block functions (i.e., \OM, \LO, \Jump$_3$, and \Eps), of which configurations correspond to F5 and F10 in Table~\ref{tab:problems}. Figure~\ref{fig:SELF-UG} and Figure~\ref{fig:SELF-DAG} illustrate the convergence in terms of the best-so-far fitness $f$ and the blocks values $v_i, i\in [4]$. 
\begin{table*}[htb]
\centering
\small
\begin{tabular}{l|r|r|r|r|r}
\toprule
 & $(1+(\lambda,\lambda))$ GA & $(1+\lambda)$ EA & $(1+\lambda)~$two-rate EA & $(1+\lambda)~$var EA & fGA \\
\midrule
\OM & 474 (4) & 284 (2) & 504 (5) & 273 (1) & 346 (3)\\
\LO & 2\,070 (5) & 970 (2) & 1542 (4) & 903 (1) & 1190 (3) \\
\Jump & 216\,926 (5) & 130\,661 (4) & 65\,110 (1) & 91\,394 (2) & 126\,513 (3)  \\
\Eps & 22\,636 (5) & 10\,241 (2) & 9\,386 (1) & 12\,756 (3) & 15\,609 (4) \\
DBP (F5) & 194\,190 (5) & 37\,086 (2) & 15\,505 (1) & 67\,061 (4) & 43\,790 (3) \\
GCP (F16) & 156\,569 (4) & 46\,326 (2) & 62\,190 (3)& 175\,894 (5) & 40\,883 (1) \\
\bottomrule
\end{tabular}
\caption{FEs and ranks (in brackets) of the EAs to reach the optimum of  $40$D problems. Results are the average of $50$ runs.}
\label{tab:SELF}
\end{table*}

As shown in Figure~\ref{fig:SELF-UG}, for the DBP where improvements in block functions simultaneously contribute to the objective $f$, the values of all block functions are optimized from the early stage of optimization. In contrast, due to the presence of gate constraints in GCP, the four block values are optimized sequentially along the optimization process, as shown in Figure~\ref{fig:SELF-DAG}-b. The proposed benchmark preserves information at the block level, enabling distinct optimization behaviors to be directly observed and analyzed. 

Table~\ref{tab:SELF} lists the number of function evaluations (FEs) of the EAs used to achieve the optimum, and it shows that the EAs exhibit different performance across four classic problems. In particular, \var~outperforms the others on \OM~and \LO, and \tworate~shows superior performance on \Jump~and \Eps. However, when heterogeneous landscape properties of these four problems are embedded within a single search space, \tworate~retains its advantages on the DBP, but the performance of \var~deteriorates substantially on both DBP and GCP. Notably, fGA, despite its comparatively weak performance on the classic problems, achieves the best performance on the conditioned space of the GCP. This finding highlights that self-adaptive methods can behave differently when multiple landscape properties coexist.

\subsection{Diversity Helps beyond Simple Multimodal Landscapes}
In this section, we investigate a recent study on maintaining diversity to improve the performance of GAs~\cite{ren2024maintaining}. In the $(\mu+1)$ GA, of which details are available in~\cite{ren2024maintaining} and the Appendix, the solution with the worst objective value is removed from the population, with broken ties uniformly at random. The work of~\cite{ren2024maintaining} proposed a diversity maintenance mechanism that considers the solution representation in the search space. In particular, the mechanism ensures that the two solutions with the largest Hamming distance in the population are preserved. While the advantage of this approach has been theoretically proved, its empirical evaluation has been limited to the \Jump~problem. In this section, we investigate whether such a mechanism of \emph{maintaining diversity guarantees advantages beyond the simple multimodal landscape of \Jump.}

\begin{figure}[tb!]
    \subcaptionbox{DBP}{\includegraphics[trim=8 5 5 5, clip,width=0.25\linewidth]{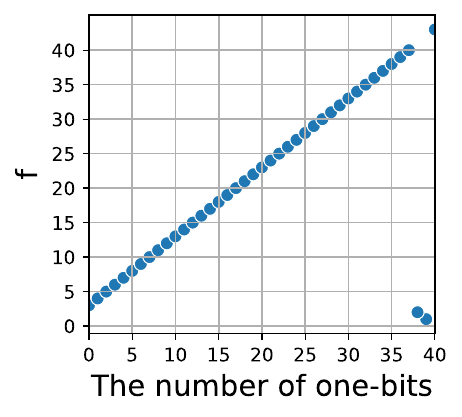}
    \includegraphics[trim=8 5 5 5, clip,width=0.25\linewidth]{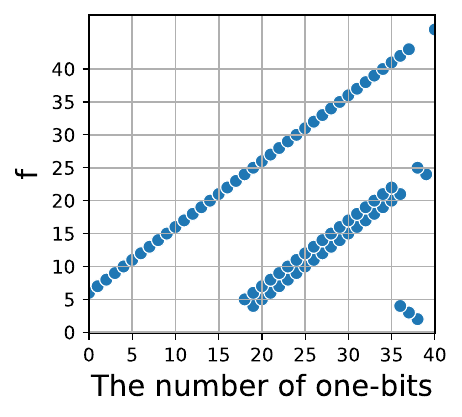}
    \includegraphics[trim=8 5 5 5, clip,width=0.25\linewidth]{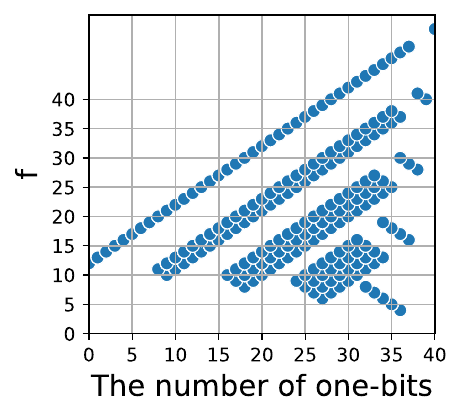}}
    \subcaptionbox{GCP}{\includegraphics[trim=8 5 5 5, clip,width=0.25\linewidth]{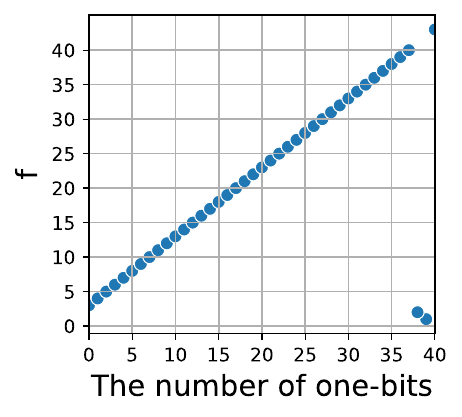}
    \includegraphics[trim=8 5 5 5, clip,width=0.25\linewidth]{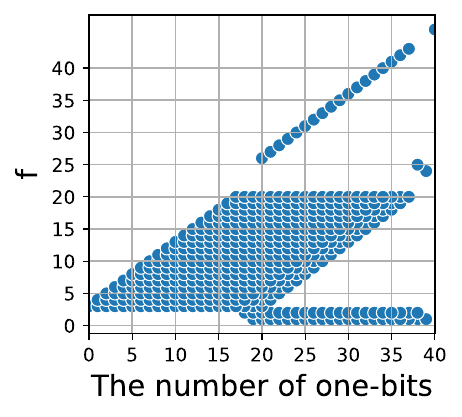}
    \includegraphics[trim=8 5 5 5, clip,width=0.25\linewidth]{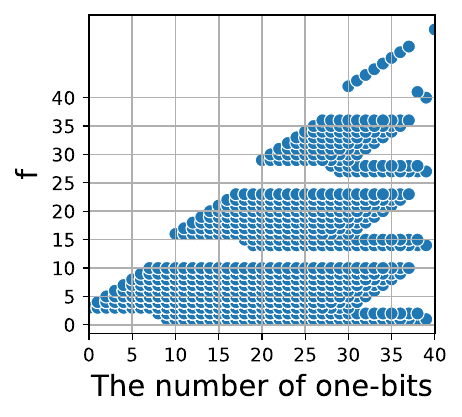}}
    \caption{The attainable objective values of a $40$D DBP (F3) and GCP (F7) with only \textsc{Jump}$_3$ block functions vs. the number of 1-bits in the solution. $m
    = 1,2,4$ from left to right.}
    \label{fig:Jump}
\end{figure}

First, we illustrate how our benchmark approach can generate novel and structured problems. Figure~\ref{fig:Jump} demonstrates the attainable objective values of the $40$D DBP (F3) and GCP (F7) constructed solely from \Jump~block functions, regarding the number of 1-bits in the solution. The problem is identical to the classic \Jump~problem when $m=1$, exhibiting a single fitness valley. As $m$ increases, the interaction among \Jump~block functions produces several disjoint regions of attainable objective values aligned along the number of 1-bits, resulting in a multimodal search space with multiple separable local optima. We can also observe that, for GCP, the attainable objective values are more densely distributed regarding the number of 1-bits, and this increased density arises from the gate constraints.

\begin{figure}[tb!]
    \subcaptionbox{\Jump~blocks}{\includegraphics[trim=8 5 5 5, clip,width=0.60\linewidth]{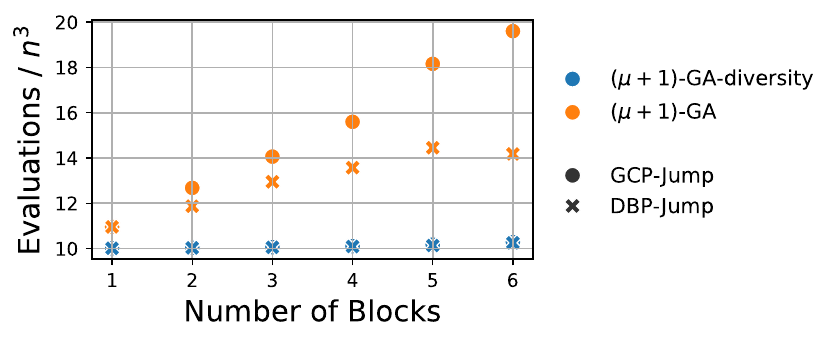}}
    \subcaptionbox{\Eps~blocks}{\includegraphics[trim=8 5 5 5, clip,width=0.40\linewidth]{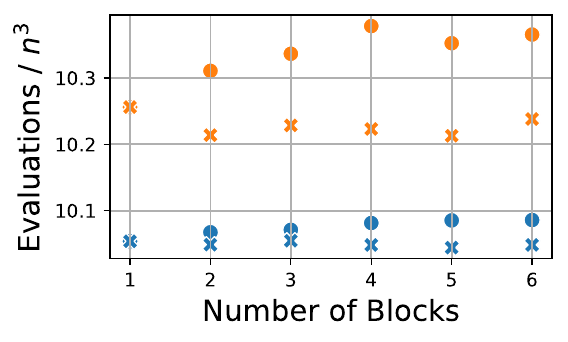}}
    \subcaptionbox{Distinct block functions}{\includegraphics[trim=8 5 5 5, clip,width=0.40\linewidth]{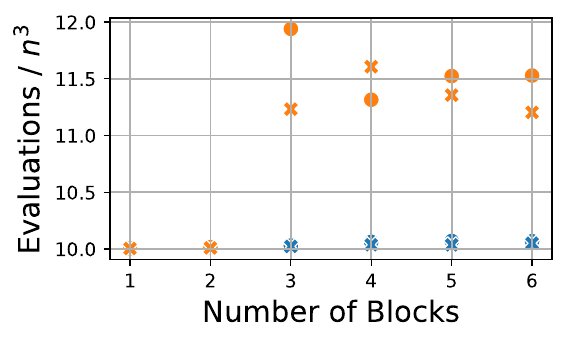}}
    \caption{The FEs of $(\mu+1)$ GAs to reach the optimum of $40$D DBPs and GCPs with \textsc{Jump}$_3$, \Eps~, and heterogeneous block functions.}
    \label{fig:MU}
\end{figure}

Figure~\ref{fig:MU} shows the FEs required by the $(\mu+1)$ GA with and without diversity maintenance to reach the optimum of $40$D problems with different block function choices. Figure~\ref{fig:MU}-a shows that, for the problem constructed solely from \Jump, the FEs required by the standard GA significantly increase substantially as $m$ increases, that is, when more multimodal structures are introduced. In contrast, the GA with diversity maintenance (GA-diversity) exhibits stable performance across different $m$. Furthermore, we examine problems constructed only from \Eps~blocks (F4 and F8, Figure~\ref{fig:MU}-b) and problems with heterogeneous block functions (F5 and F10, Figure~\ref{fig:MU}-c). Across all tested problem instances, GA-diversity consistently outperforms the standard GA. However, for the search space with only \Eps, the FEs required by GAs remain relatively stable compared to the performance shift on \Jump~blocks, as \Eps~introduces ruggedness rather than modality. For search spaces constructed by heterogeneous block functions, the advantage of GA-diversity can be observed once \Jump~and \Eps~are added into block functions ($m \ge 3$). In addition, when comparing the performance on DBP and GCP instances, DBP generally appears to be easier for both algorithms, except for a noticeable performance shift of the standard GA in Figure~\ref{fig:MU}-c at $m=4$.

Overall, our results validate that maintaining diversity can improve the performance and robustness of the GA across complex search spaces. The proposed block-structure benchmark enables systematic investigations across distinct landscapes, providing a testbed for future studies of adaptive diversity mechanisms that explicitly consider the structure of the solution representation.

\subsection{Structured Bi-objective Search Spaces}
Existing theoretical studies of EC in multi-objective optimization are largely based on classic bi-objective problems such as \OMM, \textsc{LOTZ}, etc. Recent attempts~\cite{liang2025problem} to expand these benchmarks have yielded only limited combinations of existing problem objectives. In contrast, with the ability to systematically construct objectives with transparent properties, our approach can also serve bi-objective PBO benchmarks, while the problems proposed in~\cite{liang2025problem} can be instantiated as specific cases within our framework.

\begin{figure}[tb]
    {\includegraphics[trim=8 5 5 5, clip,width=0.29\linewidth]{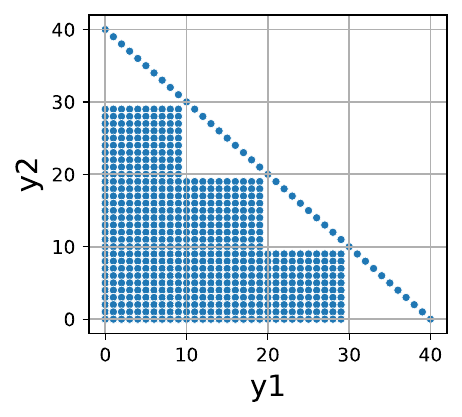}}
    {\includegraphics[trim=8 5 5 5, clip,width=0.29\linewidth]{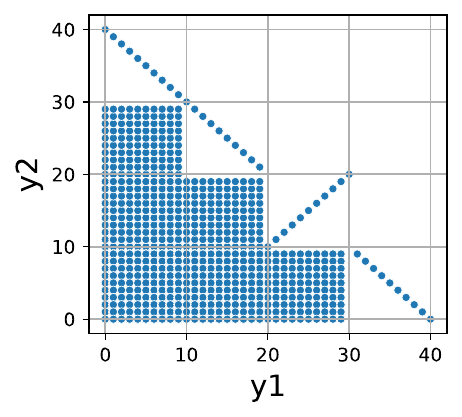}}
    \caption{The bi-objective search space of GCPs with the weight vectors of the objectives are $\{1\}^4$ vs. $\{-1\}^4$ on the \emph{Left}, and $\{1,-1,1,1\}$ vs. $\{-1,1,1,-1\}$ on the \emph{Right}.}
    \label{fig:COMP-BI}
\end{figure}

In this section, we demonstrate how different objective search spaces can be constructed by controlling the weight vector of GCPs. Figure~\ref{fig:COMP-BI} presents the objective spaces of two bi-objective optimization problems derived from $40$D GCP problem instances with four blocks $m=4$. Specifically, the two objectives $y1$ and $y2$ are defined by GCPs with gate constraint structures $E = \left\{
\begin{array}{ll}
e_{ij} = 1, &\text{if } j = i+1 \\
0, & \text{otherwise}
\end{array}
\right.$ and  $E = \left\{
\begin{array}{ll}
e_{ij} = 1, &\text{if } j = i-1 \\
0, & \text{otherwise}
\end{array}
\right.$, respectively. The corresponding parameters $A, W, \text{ and } B$ follows the relations $a_i = (1-w_i)n_i/2$ and $b_i =  (1+w_i)n_i/2, \forall i \in [4]$. We use \OM~and \LO~as the block functions in this section. Consequently, $n_i$ is the optimal value for each block $v_i$. The left objective space in Figure~\ref{fig:COMP-BI} exhibits a continuous Pareto front, indicating a strong and uniform conflict between the two objectives, corresponding to the perfectly negatively correlated weight vectors $\{1\}^m$ and $\{-1\}^m$. In contrast, under a particular setting of weight vectors $W = \{1,-1,1,1\}$ and $\{-1,1,1,-1\}$, the right objective space shows an irregular structure with a disconnected Pareto front.

\begin{figure}[tb]
    \subcaptionbox{\OM~blocks}{\includegraphics[trim=8 5 5 5, clip,width=0.3\linewidth]{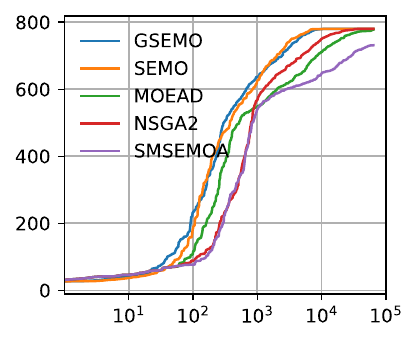}
    \includegraphics[trim=8 5 5 5, clip,width=0.3\linewidth]{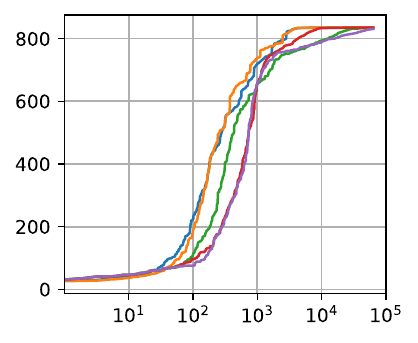}}
    \subcaptionbox{\LO~blocks}{\includegraphics[trim=8 5 5 5, clip,width=0.3\linewidth]{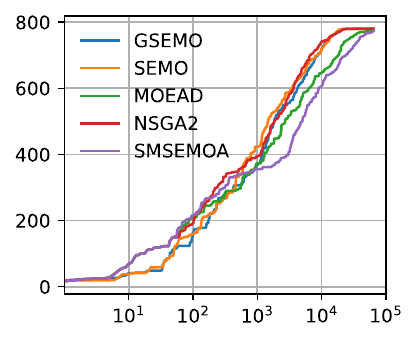}
    \includegraphics[trim=8 5 5 5, clip,width=0.3\linewidth]{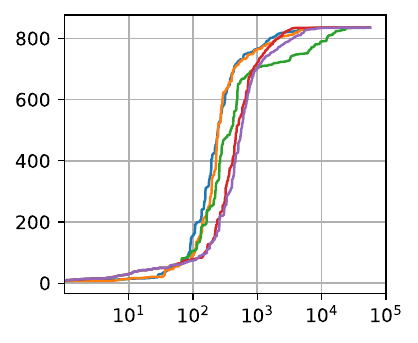}}
    \caption{The HV convergence on the bi-objective optimization problems with objective space in Figure~\ref{fig:COMP-BI}. The block functions are all \OM~in (a) and \LO~in (b).}
    \label{fig:PER-BI}
\end{figure}

We investigated the performance of five multi-objective algorithms on the four newly constructed bi-objective pseudo-Boolean problems (BF2-5). Figure~\ref{fig:PER-BI} compares the convergence in terms of Hypervolume (HV) of these algorithms. The left subfigures show the results of the GCPs with $W=\{1\}^m$ vs. $\{-1\}^m$, while the right figures correspond to another weight vector setting. The plots show the obtained HV values over the FEs, and higher convergence curves therefore indicate better performance.
For the GCPs with objectives space of weight vectors $\{1\}^4$ vs. $\{-1\}^4$, \SMS~converges competitively early on but slows down later in solving the GCP with \OM~blocks (Figure~\ref{fig:PER-BI}-a), resulting in a significant performance gap compared to the other algorithms, although \OM~is known to be easier for EAs than \LO. A similar deterioration in performance in the late stage is also observed for \SMS~when dealing with another objective space with \OM~blocks. 
For the GCPs with the disjoint Pareto front (right subfigures), GSEMO, SEMO, and \NSGA~exhibit similar performance across both \OM~and \LO~blocks, whereas \MOEAD~converges significantly more slowly than the other algorithms.

Overall, these results demonstrate that the compared algorithms are significantly affected by the objective space, which is controlled by the weight vectors. Moreover, even when targeting Pareto fronts of the same shape, the behavior of algorithms may differ greatly when the underlying search spaces in terms of solution representations, which are defined here by \OM~and \textsc{LeadingOnes}. These observations further highlight the potential of our block-structured benchmarks for systematically evaluating algorithms for general bi- and multi-objective discrete optimization.

\section{Extensions Beyond the Discussed Instances}
Beyond the concrete problem instances described in the study cases and listed in Table~\ref{tab:problems}, the proposed block-structured approach provides a general framework for systematically generating a wider range of useful discrete optimization benchmarks.

Particularly, this paper considers weight vectors $W \in \{1,-1\}^m$, associated with constant factors $A$ and constraint bounds $B$ that ensure that the objective of each block is to reach either the maximal or minimal values of the corresponding function. Regarding the adjacency matrix $E$, we consider $\{0\}^{m \times m}$ for DBPs, and graphs consisting of a single directed path traversing all blocks for GCPs. Under these settings, the proposed instances can already cover several classic benchmark problems and revisit existing studies.

These settings can be naturally extended to more general scenarios. For example, the weight factor can be generalized to $W \in \mathds{R}^m$, allowing block values to positively or negatively contribute to the objective function in different scales. Also, the dependency structure encoded by $E$ can be varied. For DBPs, dependencies between two blocks $v_i$ and $v_j$ may be defined based on their relative positions, e.g., $e_{ij} = |j - i|$. For GCPs, beyond the gate constraints defined by a single directed path, more complex and practical constraints can be formulated. For example, block functions may be assigned to multiple layers $L_l, l>2$ such that block value contributions to the objective functions are subject to hierarchical constraints. These extensions have already been included in our repository.

Regarding the bi-objective optimization, classical benchmark problems, including \OMM~(BF1) and \textsc{LOTZ}, can be naturally instantiated by our approach. For problems such as \textsc{OJZJ} and \textsc{ORZR} that are defined based on the total number of one-bits and zero-bits, direct block representation is less straightforward. However, by introducing a bijection mapping $\pi: \{0,1\}^m \rightarrow \{0,1\}^m$, the variable presentation can be transformed such that the counting ones and zeros can be embedded into the internal structure of a single block function. Under such transformation, our approach can instantiate bi-objective problems such as \textsc{OJZJ} and \textsc{ORZR} by using a single block function set via $v(\pi(x^i)), i \in [m]$.

Overall, the proposed approach enables generating a wide range of discrete benchmark problems, from covering existing classic problems to constructing more diverse instances with controllable landscape structures. We will investigate theoretical needs and practical applicability in benchmarking and further enrich the testbed for the discrete optimization community.

\section{Conclusions}
Motivated by the analysis of algorithms that address the search space of solution representations, which is a common challenge in large-scale discrete optimization, we propose a block-structure approach to construct discrete benchmarks with controllable, transparent structural landscapes. We illustrate, with several representative novel problem instances, how the approach enables systematic investigation of fundamental research questions, and establish that it can cover existing benchmarks. In particular, it facilitates the studies of (1) the self-adaptation on search spaces with heterogeneous landscape properties, (2) the practical impact of diversity mechanisms beyond simple multimodal landscapes, and (3) the behavior of multi-objective optimization across search spaces with diverse objective distributions in terms of solution representations.

Apart from the representative problem instances mentioned in this paper, we also provide a wider range of benchmarks in our repository for research exploration and communication. 
In the long term, we aim to deliver a well-structured, discrete benchmark set with diverse and explainable landscape properties.
Moreover, we expect that such a benchmark set can connect and benefit theoretical analysis and practical needs in discrete optimization. 
Beyond algorithm analysis, the proposed benchmark can also provide a foundation for advancing research on dynamic algorithm configuration and other learning-based techniques for discrete optimization.
\newpage

\section{Acknowledgments}
This publication is supported by the XAIPre project (with project number 19455) of the research program Smart Industry 2020, which is (partly) financed by the Dutch Research Council (NWO).
\bibliography{acmart}
\bibliographystyle{acm}

\clearpage
\appendix

\section{The EAs with Self-adaptive Mutation Rates}

In this section, we provide the pseudocode for the compared EAs.

The $(1+\lambda)$~EA is presented in Algorithm~\ref{alg:oplres}, and the compared EAs with self-adaptive mutation rate will focus on adjusting $p$ at line 4. The algorithm follows a standard bit mutation, which samples $\ell$ following a conditioned Binomial distribution based on a static mutation rate $p$, and then flips $\ell$ distinct bits of $x$ (i.e., $\text{Mut}_{\ell}(x)$). The conditioned sampling ensures that a $\ell > 0$ is obtained by repeating the sampling.
\begin{algorithm2e}[h]
	\textbf{Initialization:} 
	Sample $x \in \{0,1\}^{n}$ u.a.r.\;
  	\textbf{Optimization:}
	\For{$t=1,2,3,\ldots$}{
		\For{$i=1,\ldots,\lambda$}{
			Sample $\ell^{(i)}$ from $\text{Bin}_{>0}(n,p)$\;
			$y^{(i)} \leftarrow \text{Mut}_{\ell^{(i)}}(x)$\;
		}
		\label{line:selectionopl}
		Sample $x$ from $\arg\max\{f(x),f(y^{(1)}), \ldots, f(y^{(\lambda)})\}$ uniformly at random (u.a.r.)\;	
	}
\caption{The $(1+1)$~EA with a fixed mutation rate $p \in (0,1)$ for the maximization of $f$.}
\label{alg:oplres}
\end{algorithm2e}

The fGA samples $\ell$ from a static long-tail distribution. Specifically, it alters the line 4 in Algorithm~\ref{alg:oplres} by sampling $\ell$ from a power-law distribution $D_{n/2}^{\beta}$, which assigns to each  integer $k \in [n/2]$ a probability of $\Pr[D_{n/2}^{\beta}=k]={(C_{n/2}^{\beta})}^{-1}k^{-\beta}$, where $C_{n/2}^{\beta} = \sum_{i=1}^{n/2}i^{-\beta}$. We set $\beta = 1.5$ in our experiments following the suggestion in~\cite{fga}.

The \tworate~adjust a parameter $r$ online that determines the value of mutation rates, as presented in Algorithm~\ref{alg:DoerrGWY17}. In each iteration, the \tworate~creates $\lambda/2$ offspring by standard bit mutation with mutation rate $r/(2n)$, and it creates the other half of offspring with mutation rate $2r/n$. After each iteration, with probability $1/2$, $r$ is set by the value of the best offspring that been created with (ties broken u.a.r.). Otherwise, it is updated by either $r/2$ or $2r$ with the same probability. $r$ is capped by $[2,n/4]$. We use $r=2$ as the initial value in our experiment following~\cite{tworate}.

\begin{algorithm2e}[h]
	\textbf{Initialization:} 
	Sample $x \in \{0,1\}^{n}$ u.a.r and evaluate $f(x)$\;
	$r \leftarrow 2$\;
 	\textbf{Optimization:}
	\For{$t=1,2,3,\ldots$}{
		\For{$i=1,\ldots,\lambda/2$}{
			Sample $\ell^{(i)} \sim \text{Bin}_{>0}(n,r/(2n))$, 
			create $y^{(i)} \leftarrow \text{Mut}_{\ell^{(i)}}(x)$, and
			evaluate $f(y^{(i)})$\;
		 }
		\For{$i=\lambda/2+1,\ldots,\lambda$}{
			Sample $\ell^{(i)} \sim \text{Bin}_{>0}(n,2r/n)$, 
			create $y^{(i)} \leftarrow \text{Mut}_{\ell^{(i)}}(x)$, and
			evaluate $f(y^{(i)})$\;
		 }
		$x^* \leftarrow \arg\max\{f(y^{(1)}), \ldots, f(y^{(\lambda)})\}$ (ties broken u.a.r.)\;
		\If{$f(x^*)\ge f(x)$}{$x \leftarrow x^*$}
		\eIf{$x^{*}$ has been created with mutation rate $r/2$}{$s \leftarrow 3/4$}{$s \leftarrow 1/4$}
		Sample $q \in [0,1]$ u.a.r.\;
		\eIf{$q\le s$}{$r \leftarrow \max\{r/2,2\}$}{$r \leftarrow \min\{2r,n/4\}$}
	}
\caption{The two-rate $(1+\lambda)$~EA}
\label{alg:DoerrGWY17}
\end{algorithm2e}  

Unlike the $(1+\lambda)$~EA and \tworate~sampling $\ell$ from a Binomial distribution, the \var~samples $\ell$ from a normal distribution such that it can adapt not only to the expected value of $\ell$ but also its variance. The normalized bit mutation is presented at line 6 in Algorithm~\ref{alg:normalada}. We reset $c = 0$ if there exists no improvement after $n$ function evaluations and set $F=0.98$ following~\cite{varEA}.
\begin{algorithm2e}[h]%
	\textbf{Initialization:} 
	Sample $x \in \{0,1\}^{n}$ u.a.r and evaluate $f(x)$\;
	$r \leftarrow 2$\;
	$c \leftarrow 0$\;
  \textbf{Optimization:}
	\For{$t=1,2,3,\ldots$}{
		\For{$i=1,\ldots,\lambda$}{
			Sample 
			$\ell^{(i)} \sim \min\{N_{>0}(r,F^c r(1-r/n)),n\}$, 
			create $y^{(i)} \leftarrow \text{Mut}_{\ell^{(i)}}(x)$, and
			evaluate $f(y^{(i)})$\;
		 }
		$i \leftarrow \min\left\{ j \mid f(y^{(j)}) = \max\{f(y^{(k)}) \mid k \in [n]\} \right\}$\;
		\eIf{$r =\ell^{(i)}$}{$c \leftarrow c+1$}{$c \leftarrow 0$}
		$r \leftarrow \ell^{(i)}$\; 
		\If{$f(y^{(i)})\ge f(x)$}{$x \leftarrow y^{(i)}$}
	}
\caption{The \var.}
\label{alg:normalada}
\end{algorithm2e}

Furthermore, we also consider the \llGA, which applies crossover and mutation controlled by an adaptive population size $\lambda$, as presented in Algorithm~\ref{alg:ga}. Its mutation operator follows the same procedure as the algorithms introduced above, and the crossover $\text{Cross}_c(x,x^*)$ replaces $\ell_c$ distinct bits of $x$ by the values at the corresponding positions of $x^*$. $\ell_c$ is sampled by Bin$_{>0}(n,c)$. We set $F=1.5$ following the suggestion in~\cite{doerr2018optimal}.

\begin{algorithm2e}[h]
\caption{The self-adjusting $(1 + (\lambda,\lambda))$~GA}
\label{sec:ga}

\textbf{Initialization:}
Sample $x \in \{0,1\}^{n}$ u.a.r. and evaluate $f(x)$\;

\textbf{Optimization:}
\For{$t=1,2,3,\ldots$}{

\textbf{Mutation phase:\\}
Sample $\ell \sim \text{Bin}_{>0}(n,\lambda/n)$\;
\For{$i=1,\ldots,\lambda$}{
    create $y^{(i)} \leftarrow \text{Mut}_{\ell}(x)$, and evaluate $f(y^{(i)})$\;
}
$x^* \leftarrow \arg\max\{f(y^{(1)}), \ldots, f(y^{(\lambda)})\}$ (ties broken u.a.r.)\;

\textbf{Crossover phase:\\}
\For{$i=1,\ldots,\lambda$}{
    create $y^{(i)} \leftarrow \text{Cross}_c(x,x^*)$, and evaluate $f(y^{(i)})$\;
}
$y^* \leftarrow \arg\max\{f(y^{(1)}), \ldots, f(y^{(\lambda)})\}$ (ties broken u.a.r.)\;

\textbf{Selection phase:\\}
\If{$f(y^*) > f(x)$}{
    $x \leftarrow y^*$\;
    $\lambda \leftarrow \max\{\lambda/F,1\}$\;
}
\If{$f(y^*) = f(x)$}{
    $x \leftarrow y^*$\;
    $\lambda \leftarrow \min\{\lambda F^{1/4},n\}$\;
}
\If{$f(y^*) < f(x)$}{
    $\lambda \leftarrow \min\{\lambda F^{1/4},n\}$\;
}

}
\end{algorithm2e}

\section{Evolutionary Multi-objective Algorithms}
We refer the readers to~\cite{deb2016multi} for the essential concepts of multi-objective optimization and provide the definition of the performance indicator Hypervolume (HV) here:\cite{ZitzlerT99}: The hypervolume indicator $I_H(P)$ is the volume of the objective space 
that is dominated by the solution set $P$. Given a reference point $s \in \mathbb{R}^m$, $I_H(P) = \Lambda \left( \bigcup_{x \in P}[f_1(x), s_1] \times \ldots \times [f_m(x), s_m] \right)$, where $\Lambda (P)$ is the Lebesgue measure of $P$ and $[f_1(x), s_1] \times \ldots \times [f_m(x), s_m]$ is the orthotope with $f(x)$ and $r$ in opposite corners. The paper calculates HV regarding the reference point $(0,0)$.

In the following, we provide the pseudocode for the compared multi-objective evolutionary algorithms.

The Simple Evolutionary Multi-objective Evolutionary Algorithm (SEMO)~\cite{antipov2023rigorous} maintains a population $P$ of non-dominated solutions. It primarily applies mutation operators to generate new solutions and updates the Pareto front iteratively. SEMO starts with a randomly generated solution. At each iteration, a parent solution $x$ is selected from $P$ u.a.r. to generate offspring $y$ by flipping \emph{one} randomly selected bit of $x$. If $y$ dominates any solutions in $P$, those solutions will be removed from $P$, and $y$ will be added to $P$. The algorithm runs until a predefined computational budget, e.g., the maximum function evaluations, is exhausted.

The key difference between the Global SEMO (GSEMO)~\cite{antipov2023rigorous} and SEMO lies in the mutation step. 
GSEMO applies the \emph{standard bit mutation} rather than flipping exactly \emph{one bit} when creating offspring. In practice, as shown in Algorithm~\ref{alg:GSEMO}, GSEMO flips $\ell$ randomly selected bits of $x$ to create offspring $y$, and $\ell$ is sampled from a conditional binomial distribution $\text{Bin}_{>0}(n,p)$.

\begin{algorithm2e}[h]
\caption{Global SEMO}
\label{alg:GSEMO}
\textbf{Initialization:} Sample $x \in \{0,1\}^n$ u.a.r., and evaluate $f(x)$\;
$P \leftarrow \{x\}$\;
\textbf{Optimization:} \While{not stop condition}{
Select $x \in P$ u.a.r.\;
Sample $\ell \sim \text{Bin}_{>0}(n,p)$, create $y \leftarrow \text{Mut}_{\ell}(x)$, and evaluate $f(y)$\;
\If{there is no $z\in P$ such that $y \preceq z$}{$P = \{z \in P \mid z \not\preceq y\} \cup \{y\}$}
}
\end{algorithm2e}

\NSGA~\cite{nsga2}, as presented in Algorithm~\ref{alg:NSGAII}, employs non-dominated sorting to rank all the solutions and partition them into different Pareto fronts. When selecting the solutions to form the parent population, it prefers solutions located at better-ranked Pareto fronts. If multiple solutions belong to the same Pareto front, \NSGA~selects those with a larger crowding distance to ensure the diversity of populations by favoring well-spread solutions. Additionally, \NSGA~usually employs elitist selection, considering both parent and offspring populations at each iteration. Detailed calculation of \emph{non-dominated sorting} and \emph{crowding distance} can be found in~\cite{nsga2}, and we apply the implementation of the pymoo package~\cite{pymoo}.

\begin{algorithm2e}[h]
\caption{Non-dominated Sorting Genetic Algorithm}
\label{alg:NSGAII}

\textbf{Initialization:} Sample an initial population $P_0 = \{x_1,\ldots,x_\mu\}$ u.a.r., where $x \in \{0,1\}^n$, and evaluate all $x \in P_0$\;
\For{$t = 1,2,3,\ldots$}{
    Generate offspring population $Q_t$ with size $\mu$ from $P_t$\;
    Evaluate all offspring in $Q_t$\;
    $R_t \leftarrow P_t \cup Q_t$\;
    Perform \emph{non-dominated sorting} of $R_t$ into fronts $F_1,F_2,\ldots$\;
    $P_{t+1} \leftarrow \emptyset$\;
    $i \leftarrow 1$\;
    \While{$|P_{t+1}| + |F_i| \le \mu$}{
        $P_{t+1} \leftarrow P_{t+1} \cup F_i$\;
        $i \leftarrow i+1$\;
    }
    Calculate the \emph{crowding distance} of each individual in $F_i$\;
    Let ${F_i}^*$ be the first $\mu-|P_{t+1}|$ in $F_i$ with the largest \emph{crowding distance} (ties broken u.a.r.)\;
    $P_{t+1} \leftarrow P_{t} \cup {F_i}^*$
}
\end{algorithm2e}

\begin{algorithm2e}[tb]
\caption{Multiobjective selection based on dominated hypervolume}
\label{alg:SMSEMOA}

\textbf{Initialization:} Sample an initial population $P_0 = \{x_1,\ldots,x_\mu\}$ u.a.r., where $x \in \{0,1\}^n$, and evaluate all $x \in P_0$\;
\For{$t = 1,2,3,\ldots$}{
    Generate offspring population $Q_t$ with size $\mu$ from $P_t$\;
    Evaluate all offspring in $Q_t$\;
    $R_t \leftarrow P_t \cup Q_t$\;
    Perform \emph{non-dominated sorting} of $R_t$ into fronts $F_1,F_2,\ldots$\;
    $P_{t+1} \leftarrow \emptyset$\;
    $i \leftarrow 1$\;
    \While{$|P_{t+1}| + |F_i| \le \mu$}{
        $P_{t+1} \leftarrow P_{t+1} \cup F_i$\;
        $i \leftarrow i+1$\;
    }
    Calculate the \emph{hypervolume contribution} of each individual in $F_i$\;
    Let ${F_i}^*$ be the first $\mu-|P_{t+1}|$ in $F_i$ with the largest \emph{hypervolume contribution} (ties broken u.a.r.)\;
    $P_{t+1} \leftarrow P_{t} \cup {F_i}^*$
}
\end{algorithm2e}

\SMS~\cite{sms}, as presented in Algorithm~\ref{alg:SMSEMOA}, deals with multi-objective problems by focusing on optimizing hypervolume. It prefers solutions that contribute most to increasing hypervolume during the optimization process. By maximizing the dominated hypervolume, \SMS~can ensure the convergence to the Pareto front while maintaining the diversity of preserved non-dominated solutions. The difference between \NSGA~and \SMS~occurs at lines 12-13 in Algorithm~\ref{alg:SMSEMOA}, where \SMS~selects the individuals with the largest hypervolume contribution in the critical front $F_i$. Detailed calculation of hypervolume contribution can be found in~\cite{sms}.

\MOEAD~\cite{moead}, as presented in Algorithm~\ref{alg:MOEAD}, uses a decomposition-based approach to solve multi-objective problems. It transforms the original multi-objective problem into a set of scalar single-objective subproblems, based on predefined reference directions. These subproblems are optimized simultaneously, with neighboring solutions collaborating to enhance convergence. The formulation of subproblems can be found in~\cite{moead}.
\newpage

\begin{algorithm2e}[tb]
\caption{The multi-objective evolutionary algorithm based on decomposition}
\label{alg:MOEAD}

\textbf{Initialization:}\;
Generate weight vectors $\{w_1,\ldots, w_\mu\}$ and construct $h_i$ subproblem according to $w_i$. We denote $B_i$ as the neighbor of the $i$-th subproblem\;
Sample initial population $P = \{x_1,\ldots,x_\mu\}$ u.a.r. and evaluate objectives and $f_i(x_i), i\in[\mu]$\;
Initialize reference point $z^* = \{z_1^*,\ldots\} \in \{0,1\}^n$\;
Initialize $P_e = \emptyset$\;
\For{$t = 1,2,3,\ldots$}{
    \For{$i = 1,\ldots,\mu$}{
        Generate offspring $x'_i$ from $x_i$\;
        Evaluate objectives $f(x'_i)$ of $x'_i$\;
        Update reference point $\mathbf{z}$\;
        For each $y \in B_i$ such that $h_i(x'_i) \le h_i(y)$, set $y=x'_i$ and $f(y) = f(x'_i)$\;
        \If{there is no $y\in P_e$ such that $x’_i \preceq y$}{$P = \{y \in P \mid y \not\preceq x'_i\} \cup \{x'_i\}$}
    }
}
\end{algorithm2e}

\section{Genetic Algorithms}
For the $(\mu+1)$~GA, we consider the version presented in our revisiting study~\cite{ren2024maintaining}, as presented in Algorithm~\ref{alg:ga}. The corresponding GA-diversity updates line 14. Instead of removing the worst individuals u.a.r., GA-diversity preserves the individuals pair with the largest Hamming distance, then removes the worst ones from the remaining population.

\begin{algorithm2e}[tb]
\textbf{Initialization:}\\
Sample an initial population $P=\{x_0,\ldots,x_\mu\}$ u.a.r., where $x \in \{0,1\}^n$\;
Evaluate all $x \in P$\;

\textbf{Optimization:}
\For{$t = 1,2,3,\ldots$}{
    Sample $x \in P$ u.a.r.\;
    \uIf{with probability $p_c$}{
        Sample $y \in P \setminus \{x\}$ u.a.r.\;
        $x' \leftarrow \text{Crossover}(x,y)$\;
    }\Else{
        $x' \leftarrow x$\;
    }
    Flip each bit of $x'$ with probability $1/n$\;
    Evaluate $f(x')$\;
    $P \leftarrow P \cup \{x'\}$\;
    Remove one individual from $\arg\min_{z \in P} f(z)$ u.a.r.\;
}
\caption{The $(\mu+1)$ Genetic Algorithm}
\label{alg:ga}
\end{algorithm2e}

\end{document}